\documentclass{article}
\usepackage[T1]{fontenc}
\usepackage{arxiv}
\usepackage[authoryear,round]{natbib}
\usepackage{newunicodechar}
\newunicodechar{–}{\textendash}
\newunicodechar{—}{\textemdash}
\newunicodechar{‘}{\textquoteleft}
\newunicodechar{’}{\textquoteright}
\newunicodechar{“}{\textquotedblleft}
\newunicodechar{”}{\textquotedblright}

\usepackage{graphicx}
\usepackage{wrapfig}
\usepackage{float}
\usepackage{microtype}
\usepackage{amsmath,amssymb,mathtools,amsthm}
\usepackage{booktabs,multirow,tabularx,makecell,adjustbox}
\usepackage{tcolorbox}
\tcbuselibrary{breakable}
\tcbset{breakable}
\usepackage{enumitem}
\usepackage{hyperref}
\usepackage{url}
\usepackage[capitalize,noabbrev]{cleveref}

\theoremstyle{plain}

\theoremstyle{definition}

\theoremstyle{remark}

\title{PiERN: Token-Level Routing for Integrating High-Precision Computation and Reasoning}
\author{\begin{minipage}{\dimexpr\textwidth-2\tabcolsep\relax}
\centering
\normalfont\normalsize
\begin{tabular*}{\linewidth}{@{\extracolsep{\fill}}cccc@{}}
\textbf{Jingyuan Fan} & \textbf{Purui Liu} & \textbf{Hengbo Xiao} & \textbf{Yuxuan Zheng} \\
\small Peking University & \small Peking University & \small Peking University & \small Peking University
\end{tabular*}\\[9pt]
\begin{tabular*}{0.80\linewidth}{@{\extracolsep{\fill}}ccc@{}}
\textbf{Jingzhao Zhang} & \textbf{Chao Lu} & \textbf{Guannan He} \\
\small Tsinghua University & \small Tsinghua University & \small Peking University
\end{tabular*}
\end{minipage}}
\date{}
\hypersetup{hidelinks,
  pdftitle={PiERN: Token-Level Routing for Integrating High-Precision Computation and Reasoning},
  pdfauthor={Jingyuan Fan, Purui Liu, Hengbo Xiao, Yuxuan Zheng, Jingzhao Zhang, Chao Lu, Guannan He}}

\begin{document}
\maketitle

\begin{abstract}
Tasks on complex systems require high-precision numerical computation to support decisions. However, current large language models (LLMs), even with enhanced reasoning capabilities, cannot integrate such computations as an intrinsic and interpretable capability with existing architectures. To this end, we propose \textbf{Physically-isolated Experts Routing Network} (PiERN), an architecture that directs computation and reasoning at token level, thereby enabling iterative alternation within a single chain of thought. We systematically evaluate PiERN on representative computation–reasoning tasks, including PDEBench and battery management tasks. Results show that PiERN achieves not only higher accuracy than directly finetuning LLMs but also significant improvements in response latency, token usage, GPU energy consumption, and experts routing accuracy compared with mainstream multi-agent approaches, while exhibiting no significant degradation in performance on MMLU and GLUE benchmarks. PiERN offers an efficient, interpretable, and scalable paradigm for interfacing language models with scientific systems.
\end{abstract}

\section{Introduction}
\begin{wrapfigure}{r}{0.50\textwidth}
    \centering
    \includegraphics[width=\linewidth]{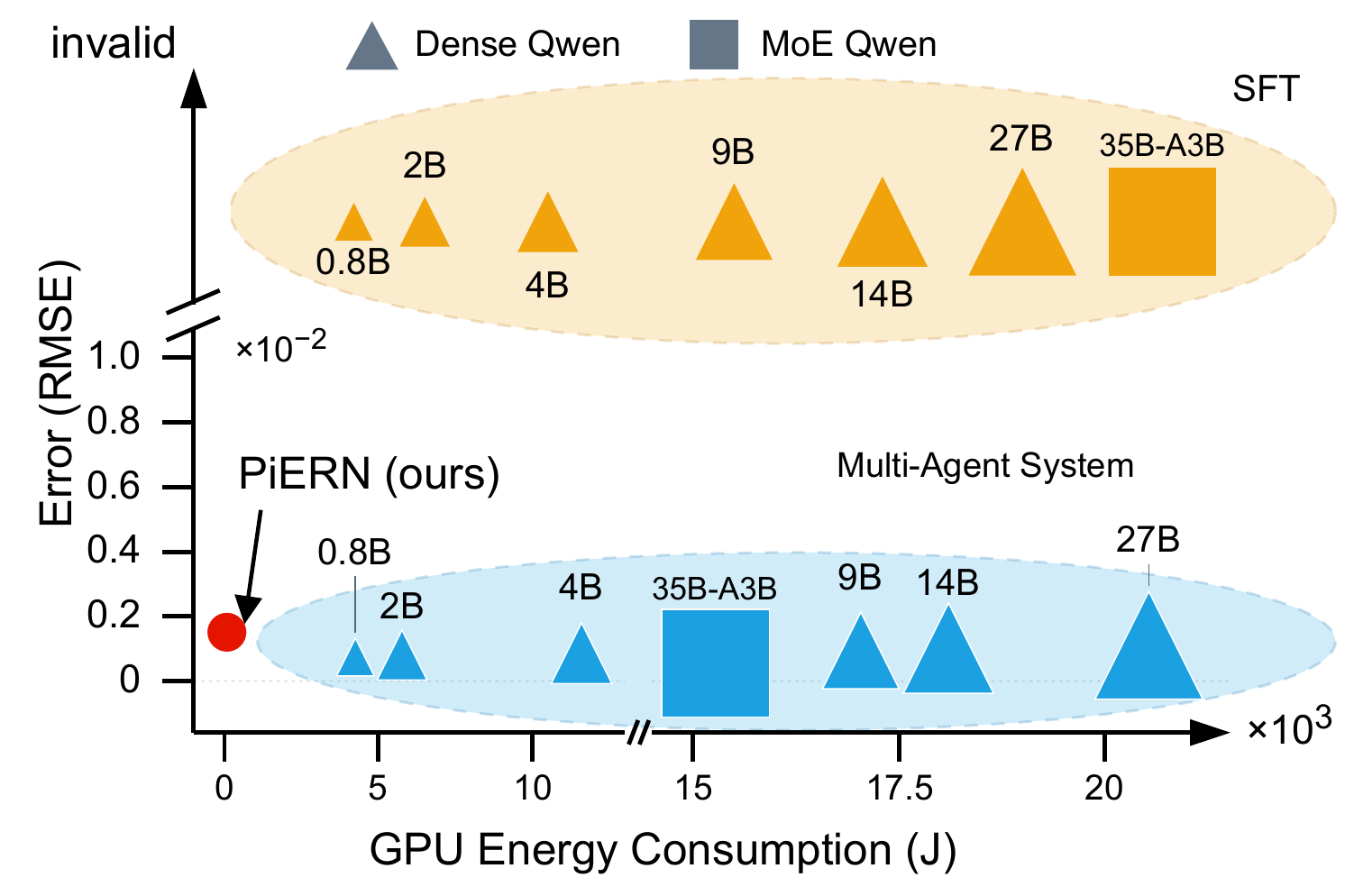}
    \caption{\textbf{PiERN achieves high precision with low GPU energy consumption.} The horizontal axis represents GPU energy consumption, and the vertical axis represents normalized RMSE on Diffusion-Sorption.}
    \label{fig:Intro}
\end{wrapfigure}

Decision-making processes in complex systems arising from scientific research and engineering practice often rely critically on high-precision numerical computation results \citep{Kennedy2022Bayesian, hennig2015probabilistic}. For example, power grid scheduling in modern power systems requires the accurate prediction of system loads and renewable energy generation to support reliable and optimal scheduling decisions \citep{HASAN2025100922, en17215385, BISWAL20243654}. Although large language models (LLMs) have recently achieved breakthrough progress in language understanding and logical reasoning, they still exhibit significant shortcomings in their intrinsic ability for high-precision numerical computation \citep{yang2025numbercookbooknumberunderstanding}. LLMs can generate seemingly reasonable chains of logic during reasoning, but once high-precision float operations, multi-step calculations, or partial differential equations (PDEs) solving are involved, they often yield incorrect or imprecise results \citep{Huang_2025, jiang2025llmsmathematicalreasoningmistakes}. This deficiency severely constrains the applicability of LLMs in complex system decision-making scenarios \citep{alampara2025probinglimitationsmultimodallanguage}. 

To compensate for this deficiency, existing studies have mainly pursued two approaches. The first approach is to perform end-to-end finetuning of LLMs, enabling them to directly learn numerical computation capabilities. However, it often remain insufficient when the
task inputs and outputs take the form of high-dimensional
numerical matrices or spatiotemporal grid solutions (e.g.,
PDE solution fields) \citep{bao2025texttrainedllmszeroshotextrapolate, 10.1093/jamia/ocae090, YANG2025107455}, and may face catastrophic forgetting \citep{li-etal-2024-revisiting, kotha2024understandingcatastrophicforgettinglanguage}. 
The second approach is based on multi-agent systems that invoke external experts, where LLMs act as the central brain \citep{Schick2023Toolformer,wu2023autogen,li2025searcho1agenticsearchenhancedlarge}, responsible for functions such as task understanding and scheduling, while external experts are responsible for executing specific high-precision computations. Although this approach ensures the accuracy, it inevitably introduces additional communication and coordination overhead, leading to low reasoning efficiency, high response latency, large GPU energy consumption \citep{chen2024optima}, and limited scalability in large-scale deployments.
\begin{figure}[t]
    \centering
    \includegraphics[width=\textwidth]{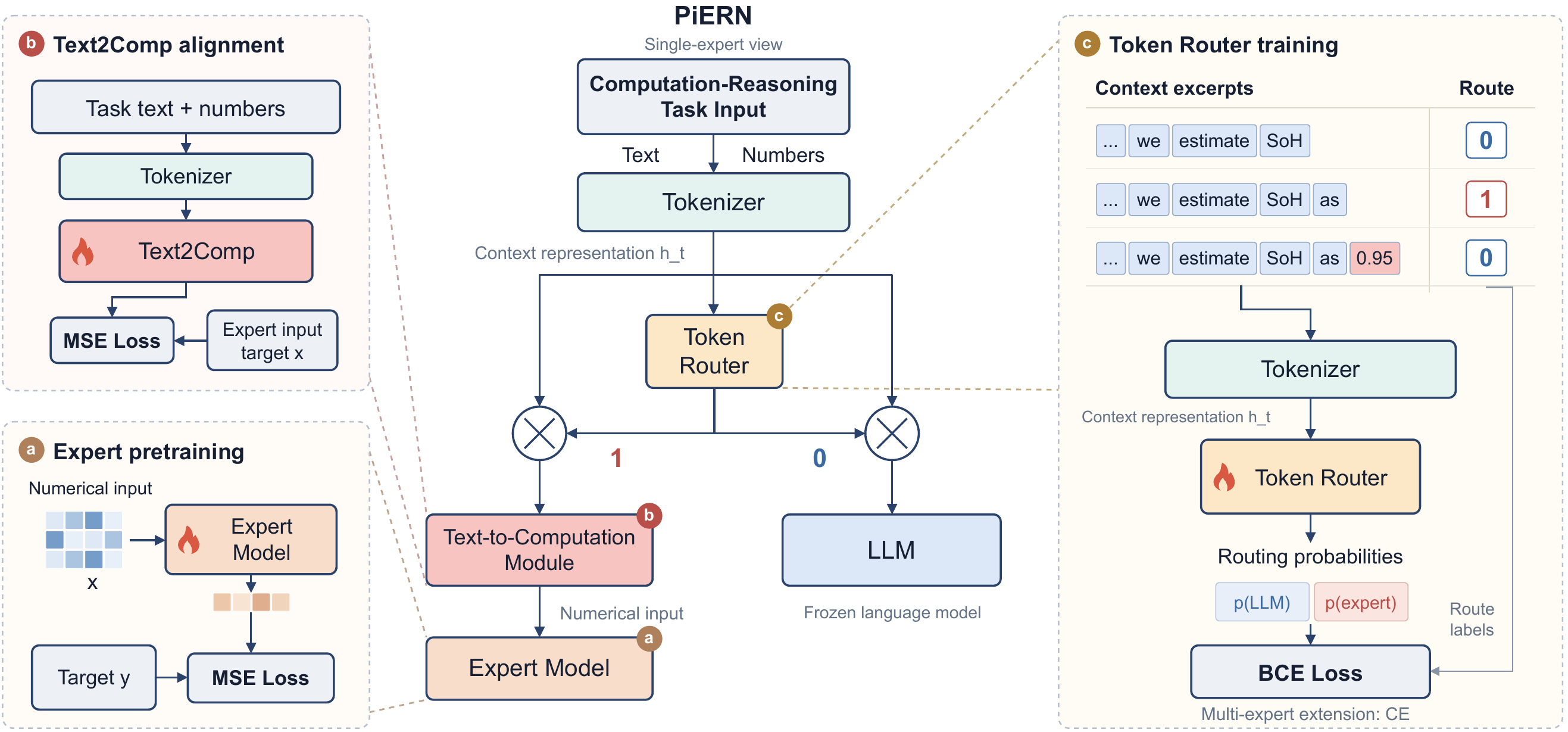}
    \caption{\textbf{(a):} Training of Expert Model for specific high-precision numeric computation tasks. \textbf{(b):} Training the Text-to-Computation Module for alignment, we adpot LLM as text-to-computation Decoder. \textbf{(c):} Training the Token Router to determine LLM for next token prediction or experts for high-precision computational results. \textbf{Middle: The overall architecture of PiERN}.}
    \label{fig:moe-train}
\end{figure}
Recent work has explored directly integrating high-precision computation capabilities into language modeling \citep{wu2024scalingparticlecollisiondata, mcleish2024transformers}. Although these approaches demonstrate promising performance on specific tasks, they still remain difficult to extend to more complex computation–reasoning scenarios such as text-to-computation and multi-step computation, due to the lack of coupling computation with reasoning at the architectural level.

Therefore, how to faithfully and efficiently integrate high-precision numerical computation with language reasoning, has become the core scientific problem in advancing next-generation scientific intelligence systems.

To address this challenge, we propose \textbf{Physically-isolated Experts Routing Network} (PiERN), an architecture that integrates high-precision scientific computation experts for complex systems state solving with the language reasoning of LLMs, regardless of whether such experts are realize via neural or non-neural computational patterns. We compare the GPU energy consumption and precision to further highlight the limitations of existing approaches in high-precision computation-reasoning tasks. As shown in Figure~\ref{fig:Intro}, when viewed along the x-axis, multi-agent systems are able to achieve relatively high-precision, but cost extremely large GPU energy consumption. Along the y-axis, the finetuned LLMs even fail to reliably follow instructions, far from attaining sufficient computational accuracy, when the inputs and outputs are high-dimensional numerical matrix. In contrast, PiERN simultaneously overcomes both limitations, achieving high-precision with minimal GPU energy consumption, thereby demonstrating clear advantages in efficiency and scalability.

As illustrated in Figure. \ref{fig:moe-train}, the core idea of PiERN is to enable dynamic switching and efficient coordination between high-precision computation experts and LLMs reasoning at the token level. In contrast to multi-agent systems through external experts invocation, PiERN adopts a fundamentally different computation–reasoning granularity by internalizing experts coordination within a single chain of thought, thereby enabling coupling between high-precision computation and reasoning with preserved interpretability and efficiency. PiERN consists of three components: physically-isolated high-precision scientific computation experts, a text-to-computation module, and a token router (Sec.~\ref{sec:PiERN Methodology}). We conducted systematic evaluations of PiERN on a range of representative computation–reasoning tasks, including PDEBench \citep{10.5555/3600270.3600387} and battery management tasks (Sec.~\ref{sec:experiments}). The results show that PiERN significantly outperforms finetuned LLMs in prediction accuracy, and multi-agent baselines in inference cost; moreover, with the language expert kept frozen, PiERN exhibits no significant degradation on the Massive Multitask Language Understanding (MMLU) \citep{MMLU-benchmark} and the General Language Understanding Evaluation (GLUE) \citep{GLUE-benchmark} benchmarks. 

Our contributions are summarized as follows:
\begin{enumerate}
    \item We introduce PiERN, an architecture that natively integrates physically-isolated high-precision scientific computation experts into LLMs, and under this architecture systematically design a stepwise training method and a token-level dynamic expert routing inference paradigm, thereby achieving the unity of accuracy, interpretability, and inference efficiency.
    \item We demonstrate that, by adopting LLM as text-to-computation decoder, PiERN learns semantically conditioned transformations that directly parameterize the numerical input values of high-precision experts, thereby fundamentally extending beyond traditional time-series and scientific patterns that are restricted to purely numerical inputs.
    \item We reveal PiERN to support reasoning-driven compositional high-precision computation. The computational results are propagated via reasoning chains and routed to other experts to enable collaborative and iterative computation, providing a new modeling paradigm for multi-stage computation-reasoning tasks.
    \item We construct and release standardized computation–reasoning task datasets, including text-conditioned extensions of PDEBench, as well as a battery-domain dataset supporting multi-expert collaborative and iterative computation, enabling reproducible and comparable evaluation of PiERN and related researches.
\end{enumerate}

\section{Related Work}
\textbf{Multimodal Capabilities of LLMs} Multimodal learning has become a central direction in today's AI, aiming to integrate text, vision, and audio within unified architectures ~\citep{zhang2024mmllmsrecentadvancesmultimodal, zhang2023metatransformerunifiedframeworkmultimodal}. This field is systematically reviewed in surveys that summarize key architectures, training strategies, and challenges \citep{wu2023multimodal}. Recent breakthroughs such as GPT-4o ~\citep{openai2024hellogpt4} and GLM-4.5V ~\citep{vteam2025glm45vglm41vthinkingversatilemultimodal} highlight these rapid progress. Mixture-of-Experts (MoE) ~\citep{jacobs1991adaptive} has emerged as an efficient paradigm, demonstrating strong performance in multimodal generation, alignment, and controllable content creation ~\citep{li2024unimoescalingunifiedmultimodal, chen2024eve, qin2023unicontrolunifieddiffusionmodel}. Despite these advances, their utility in high-precision computation remains limited ~\citep{jiang2025mmadcomprehensivebenchmarkmultimodal}.

\textbf{Mathematical and Reasoning Abilities} LLMs have achieved impressive results in mathematical reasoning, with comprehensive surveys outlining the field's progression from comprehension to complex answer generation ~\citep{wang2025survey}. To enhance the reliability of multi-step reasoning, research has focused on integrating feedback mechanisms, such as process and outcome rewards, to guide the reasoning process ~\citep{wei2025survey}. Despite these advances, LLMs still fail at improving the fundamentalnumerical understanding and processing abilities (NUPA)\citep{yang2025numbercookbooknumberunderstanding} via end-to-end finetuning. Multi-agent systems mitigate this gap by leveraging external experts through function calls ~\citep{Schick2023Toolformer, Patil2023Gorilla, wu2023autogen}, significantly enhancing high-precision capabilities, yet they still suffer from communication overhead and limited scalability ~\citep{chen2024optima}.

\textbf{Integrating High-Precision Computation with Language} Recent attempts have explored directly embedding numerical representations into models \citep{mcleish2024transformers, wu2024scalingparticlecollisiondata}, as well as adapting LLMs to structured numerical domains such as time series through reprogramming or tokenization schemes \citep{jin2023time, ansari2024chronos}. A core challenge is the mismatch between the text-generation paradigm and high-dimensional numerical tasks. Standard tokenizers fragment continuous values, undermining numerical integrity \citep{spathis2024first, zhang2024counting}, while autoregressive struggles with long, structured outputs like fine-grained PDE solutions \citep{bao2025text}. 

\section{PiERN Methodology}
\label{sec:PiERN Methodology}

In this section, we detail the methodology of PiERN. PiERN integrates high-precision scientific computation experts and LLMs as modules within the same architecture. This integration is interpretable, supporting controllable training and efficient inference. We first provide an overview of the overall PiERN architecture, followed by a stepwise training method for each component module, and finally, we present the inference paradigm of alternating invocation of different experts at the token-level granularity.

\subsection{Architecture Overview}

As shown in Figure~\ref{fig:moe-train}, the PiERN architecture consists of three core components: (i) a set of high-precision scientific computation experts, which are trained on domain-specific data; (ii) a text-to-computation module, which aligns the inputs of computation-reasoning task inputs with experts input representations; and (iii) a token router, which dynamically decides whether to invoke an expert or the LLM.

\subsection{Stepwise Training}

We propose a stepwise training method that decouples the training processes of different modules in PiERN, reduces the interference between heterogeneous optimization objectives of numerical computation and natural language.

\textbf{Stage 1: Expert Model Pre-training.}  
As shown in Figure~\ref{fig:moe-train}(a), for neural network based high-precision scientific computation experts, the first stage involves training on fixed numerical input-output pairs, where the data come from a scientific or industrial domain. It should be noted that this stage can be skipped for non-neural network experts. Let the training data be $(\mathbf{x}, \mathbf{y}) \in \mathcal{D}_{\text{exp}}$, where $\mathbf{x}$ denotes the input conditions and $\mathbf{y}$ denotes the corresponding ground-truth high-precision numerical computation results. The expert model $f_\theta$ approximates the true mapping by minimizing the mean squared error (MSE): 
\begin{equation}
\mathcal{L}_{\text{exp}} = \frac{1}{N}\sum_{i=1}^N \lVert f_{\theta}(\mathbf{x}_i) - \mathbf{y}_i \rVert^2.
\label{eq:exp}
\end{equation}
After convergence, the parameters of the expert model are frozen to maintain its high-precision scientific computation capability during subsequent PiERN inference.
\begin{figure}[t]
    \centering
    \includegraphics[width=0.80\textwidth]{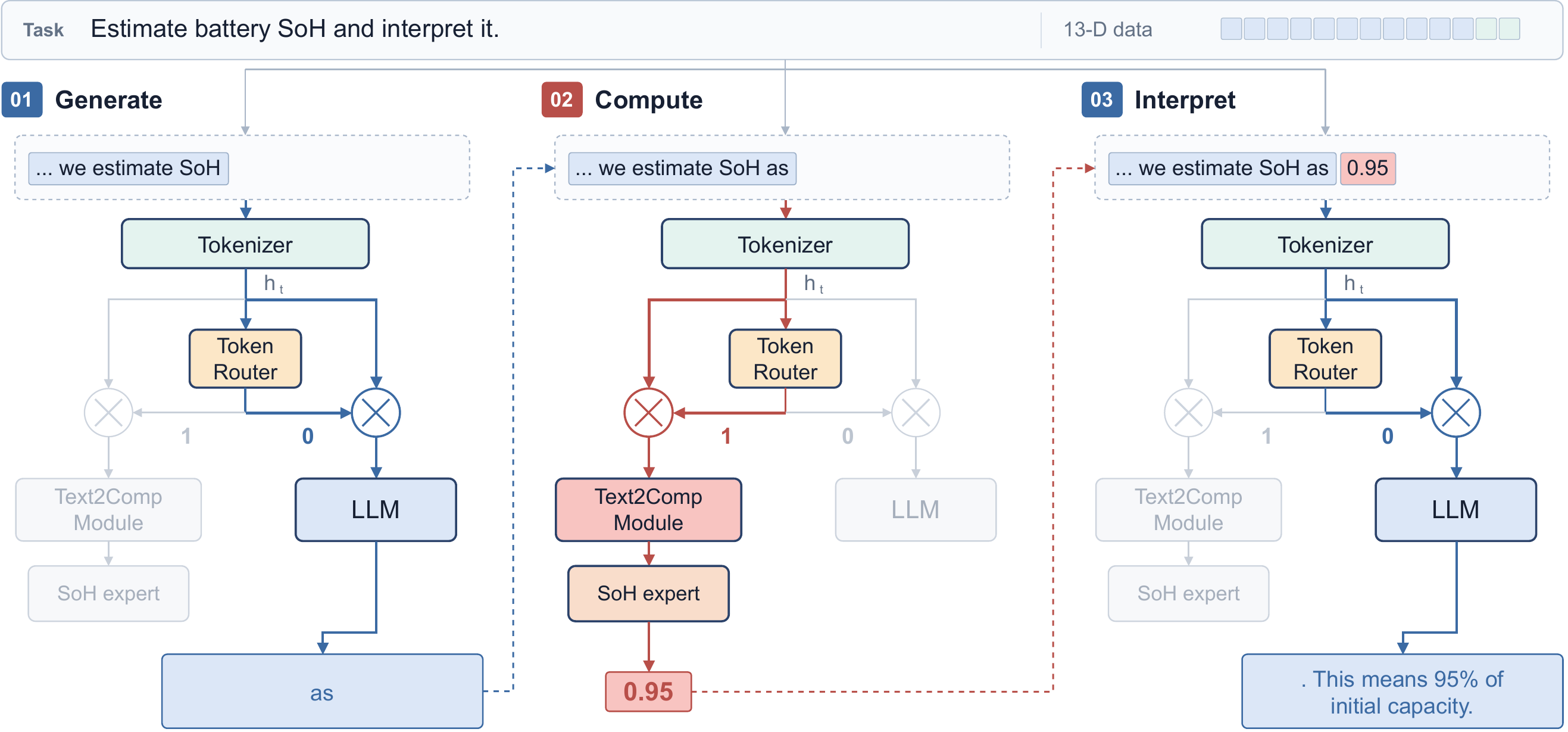}
    \caption{\textbf{Token-Level Routing for reasoning-computation inference paradigm in PiERN}. \textbf{Left:} Token Router decides to send tokenized inputs into LLM for generating the next token. \textbf{Middle:}Token Router decides to send tokenized inputs into the expert model for high-precision computation. \textbf{Right:} Token Router decides to send tokenized inputs with computation results to LLM for subsequent reasoning and planning.}
    \label{fig:moe-inference}
\end{figure}

\textbf{Stage 2: Text-to-Computation Module Training.}  
In the second stage, we optimize the text-to-computation module so that it can align inputs of language computation-reasoning task with the inputs of high-precision scientific computation experts. The training data are $(\mathbf{s}, \mathbf{x}) \in \mathcal{D}_{\text{text2comp}}$, where $\mathbf{s}$ denotes the natural language computation-reasoning task inputs and $\mathbf{x}$ denotes the structured numerical inputs required by the experts. The mapping function $g_\phi$ learns to project text inputs into numerical input representations that are not only compatible with the experts but also capture how the semantics of the text may inherently influence the appropriate inputs for the experts, as shown in Figure~\ref{fig:moe-train}(b), by minimizing the MSE loss: 
\begin{equation}
\mathcal{L}_{\text{text2comp}} = \frac{1}{N}\sum_{i=1}^N \lVert g_{\phi}(\mathbf{s}_i) - \mathbf{x}_i \rVert^2.
\label{eq:text2comp}
\end{equation}
To further strengthen the alignment between semantics and numerical values, we optionally introduce a contrastive loss~\citep{Oord2018RepresentationLW}, inspired by its successful application in cross-modal representation learning such as CLIP for vision–language alignment~\citep{Radford2021LearningTV}:
\begin{equation}
\mathcal{L}_{\text{contrastive}} = - \sum_{i=1}^N \log 
\frac{\exp(\text{sim}(g_\phi(\mathbf{s}_i), \mathbf{x}_i)/\tau)}
{\sum_{j=1}^N \exp(\text{sim}(g_\phi(\mathbf{s}_i), \mathbf{x}_j)/\tau)},
\label{eq:contrastive}
\end{equation}
where $\text{sim}(\cdot,\cdot)$ denotes the similarity function (e.g., cosine similarity), and $\tau$ denotes the temperature coefficient. The final training objective is defined as the weighted sum of ~\eqref{eq:text2comp} and ~\eqref{eq:contrastive}:
\begin{equation}
\mathcal{L}_{\text{stage2}} = \mathcal{L}_{\text{text2comp}} + \lambda \, \mathcal{L}_{\text{contrastive}},
\label{eq:stage2}
\end{equation}
where $\lambda$ is the balancing coefficient. This joint training objective can distinguish correct and incorrect language–expert pairs, improve training efficiency, and promote the text-to-computation module to accurately regress language tasks to the correct numerical inputs required by experts.

\textbf{Stage 3: Token Router Training.}  
In the final stage, we train the token router to dynamically decide at each time step whether to invoke a high-precision scientific computation expert or the LLM. Its input is the hidden representation $\mathbf{h}_t$ of all tokens at the current time step, and its output is a probability distribution $p(e \mid \mathbf{h}_t)$ over the set of experts and the LLM $\mathcal{E}$, which indicates which expert model or the LLM should be selected in the next-token prediction process. The training data are $(\mathbf{h}_t, e_t) \in \mathcal{D}_{\text{router}}$, where $e_t$ denotes the corresponding token-level invocation label, as shown in Figure~\ref{fig:moe-train}(c). The router is optimized using a cross-entropy (CE) loss:
\begin{equation}
\mathcal{L}_{\text{router}} = - \sum_{t} \sum_{e \in \mathcal{E}} y_{t,e} \log p(e \mid \mathbf{h}_t),
\label{eq:router}
\end{equation}
where $y_{t,e}$ is a one-hot vector. In particular, when $\mathcal{E}$ contains only one expert and the LLM, this objective naturally degenerates into the binary cross-entropy (BCE) loss.

\subsection{Inference Paradigm}

During the inference paradigm shown in Figure~\ref{fig:moe-inference}, PiERN integrates the pre-trained high-precision experts, text-to-computation module, token router, and the LLM via neural network connections. This integrated model is then used to execute language computation-reasoning tasks, and subsequent inference and planning are carried out based on the high-precision computation results.

Building on this architecture, PiERN dynamically switches between standard language reasoning and high-precision computation of expert model. For example, given inputs “The battery health”, the token router detects no computation requirement (as the next token is likely to be "is") and forwards the sequence to the LLM for ordinary next-token prediction. In contrast, given “The battery health is” when a concrete numeric value is required, the router invokes the text-to-computation module to transform the sequence into expert inputs; then the expert model returns a high-precision value (e.g., 0.95), which is appended to the sequence as “The battery health is 0.95.” The computed value is seamlessly incorporated into the context, enabling the LLM to continue reasoning (e.g., “, which is in a relatively good state for daily use.”). In this way, PiERN preserves numerical results accuracy while maintaining coherent language reasoning, planning, and decision-making.

\section{Experiments and Results}
\label{sec:experiments}

\subsection{Task Descriptions}
\label{sec:task-descriptions}

We evaluate PiERN on five PDE tasks: Diffusion-Reaction, Diffusion-Sorption, Burgers, Navier-Stokes, and Acoustic Wave \citep{10.5555/3600270.3600387,deng2021openfwi}. These tasks ensure a comprehensive evaluation of PiERN across a range of scenarios, from linear to nonlinear, and from simple diffusion processes to more complex interactions involving multiple physical phenomena. Further task details are provided in Appendix~\ref{app:PDEBench_Task}.

\subsection{Language Templates and Data Synthesis}
\label{sec:language-data}

For these five tasks, we construct text-numerical computation-reasoning data by embedding numerical inputs in natural-language instructions. For all five tasks, the text-to-computation data use 80 training templates and 20 unseen test templates. The token-router labels $y \in \{0, 1, 2, 3, 4, 5\}$ indicate the LLM or one of the five task-specific experts. Appendix~\ref{app:data_formulation} describes this construction. The standalone component study below uses a separate paired-template protocol to evaluate generalization independently of the system-level comparisons.

\subsection{Generalization of Core Components}
\label{sec:component-generalization}

\textbf{Experimental setup.} We evaluate the text-to-computation module (text2comp) and Token Router separately on Diffusion-Sorption using seen and unseen instruction templates. We measure input reconstruction RMSE for text2comp and decision-point routing accuracy for the Router. Detailed experimental settings are provided in Appendix~\ref{app:component-generalization}.

\begin{table}[!ht]
\centering
\caption{Generalization of core components to unseen instruction templates.}
\label{tab:component-generalization}
\small
\setlength{\tabcolsep}{4pt}
\begin{tabular*}{0.98\textwidth}{@{\extracolsep{\fill}}llcc@{}}
\toprule
\textbf{Component} & \textbf{Metric} & \textbf{Seen templates} & \textbf{Unseen templates} \\
\midrule
text2comp & Reconstruction RMSE ($\times 10^{-3}$) $\downarrow$ & 6.607 & 6.569 \\
Token Router & Routing accuracy $\uparrow$ & 100\% & 100\% \\
\bottomrule
\end{tabular*}
\end{table}

\textbf{Experimental results.} Table~\ref{tab:component-generalization} validates the generalization of both core components to unseen instruction-template combinations. text2comp achieves mean reconstruction RMSEs of $6.607\times10^{-3}$ on seen templates and $6.569\times10^{-3}$ on unseen templates. The nearly identical errors show that numerical alignment remains stable under the tested template changes, with no observed increase in mean reconstruction error on the unseen split.

The Token Router achieves $100\%$ test accuracy on both splits, with no false expert invocations or missed invocations at the evaluated decision points. This result also holds separately for Identity, Scaling, and Offset instructions. Thus, the learned decision of when to invoke computation transfers to unseen template combinations rather than being restricted to the complete templates used during training.

Together, these results experimentally validate two complementary capabilities: reconstructing numerical expert inputs and selecting computation-invocation points under unseen instruction-template combinations.

\subsection{PiERN}
\begin{wrapfigure}{R}{0.55\textwidth}
    \centering
    \includegraphics[width=\linewidth]{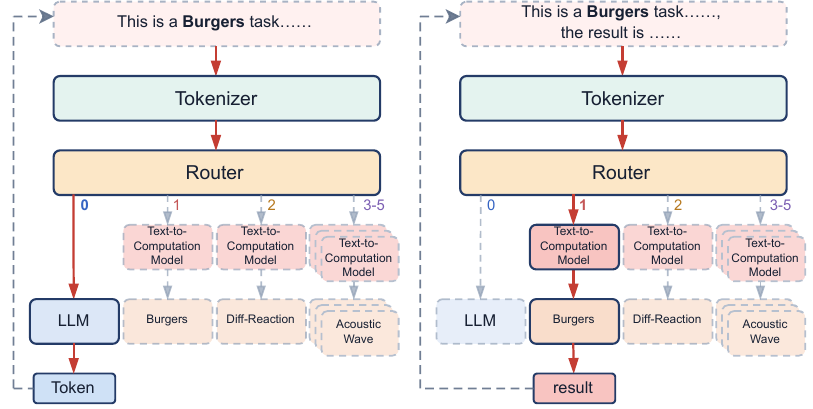}
    \caption{\textbf{A five-expert example of the PiERN architecture}. \textbf{Left:} Routing to the LLM expert for next token prediction and reasoning. \textbf{Right:} Routing to Burgers expert for high-precision computation.}
    \label{fig:router-multi-expert}
\end{wrapfigure}

Figure~\ref{fig:router-multi-expert} presents a five-expert example of the PiERN architecture. PiERN-2.4B uses Qwen3.5-0.8B as its base language model. The task-specific text2comp module and Token Router are each obtained by finetuning Qwen3.5-0.8B. For each task, these three modules have a nominal total of 2.4B parameters: 0.8B from the base language model and an additional 1.6B from text2comp and the Token Router. Other PiERN variants differ only in the base LLM used for language generation; the text2comp modules, Token Router, and overall architecture remain unchanged. The routing accuracy reaches $100\%$ during training. Simultaneously, the numerical reconstruction MSE of the text-to-computation module  demonstrates excellent alignment: training and test errors remain closely aligned, with MSE reaching $10^{-4}$. This indicates that PiERN can accurately bridge the gap between natural language instructions and formal computational inputs for experts. It is worth noting that under the PiERN architecture, with a $100\%$ experts routing success rate, the computation error primarily arises from the text-to-computation module and the expert model. Appendix~\ref{app:PDEBench end-to-end runing example} show an end-to-end running example of PiERN.

\WFclear
\subsubsection{Performance over LLM finetuning}

Finetuning is a common solution to bring domain knowledge into LLMs, thereby it can also be used to enhance LLMs with high-precision computation capabilities. Across five tasks, we compare RMSE on normalized outputs for the downsampled models, PiERN, and the fine-tuned models.
\begin{table}[t]
\caption{RMSE comparison between PiERN and base models}
\centering
\setlength{\tabcolsep}{4pt}
\begin{adjustbox}{scale=0.9}
\begin{tabular*}{\textwidth}{@{\extracolsep{\fill}}lccccc}
\toprule
\textbf{Model} & \makecell{\textbf{Diffusion}\\\textbf{Reaction}} & \makecell{\textbf{Diffusion}\\\textbf{Sorption}} & \textbf{Burgers} & \makecell{\textbf{Navier--}\\\textbf{Stokes}} & \makecell{\textbf{Acoustic}\\\textbf{Wave}} \\
\midrule
Downsampled Model & 0.000334 & 0.000488 & 0.000064 & 0.000503 & 0.042331 \\
PiERN & \textbf{0.001151} & \textbf{0.001371} & \textbf{0.000516} & \textbf{0.001382} & \textbf{0.054738} \\
Fine-tuned Model & \textbf{invalid} & \textbf{invalid} & \textbf{invalid} & \textbf{invalid} & \textbf{invalid} \\
\bottomrule
\end{tabular*}
\end{adjustbox}
\label{tab:rmse_comparison}
\end{table}
Table~\ref{tab:rmse_comparison} shows that PiERN produces numerical predictions across all five tasks, with a limited increase in absolute RMSE over the corresponding downsampled models. The fine-tuned model outputs are marked as invalid. Moreover, the training of LLMs is an end-to-end, data-driven process, leading the fine-tuned models have very low interpretability. In contrast, our PiERN method integrates the pre-trained, and frozen experts, significantly enhancing the interpretability and stability.

\subsubsection{Performance over Multi-Agent System}
\begin{figure}[!hb]
    \centering
    \includegraphics[width=\textwidth, trim={0 40pt 0 5pt}, clip]{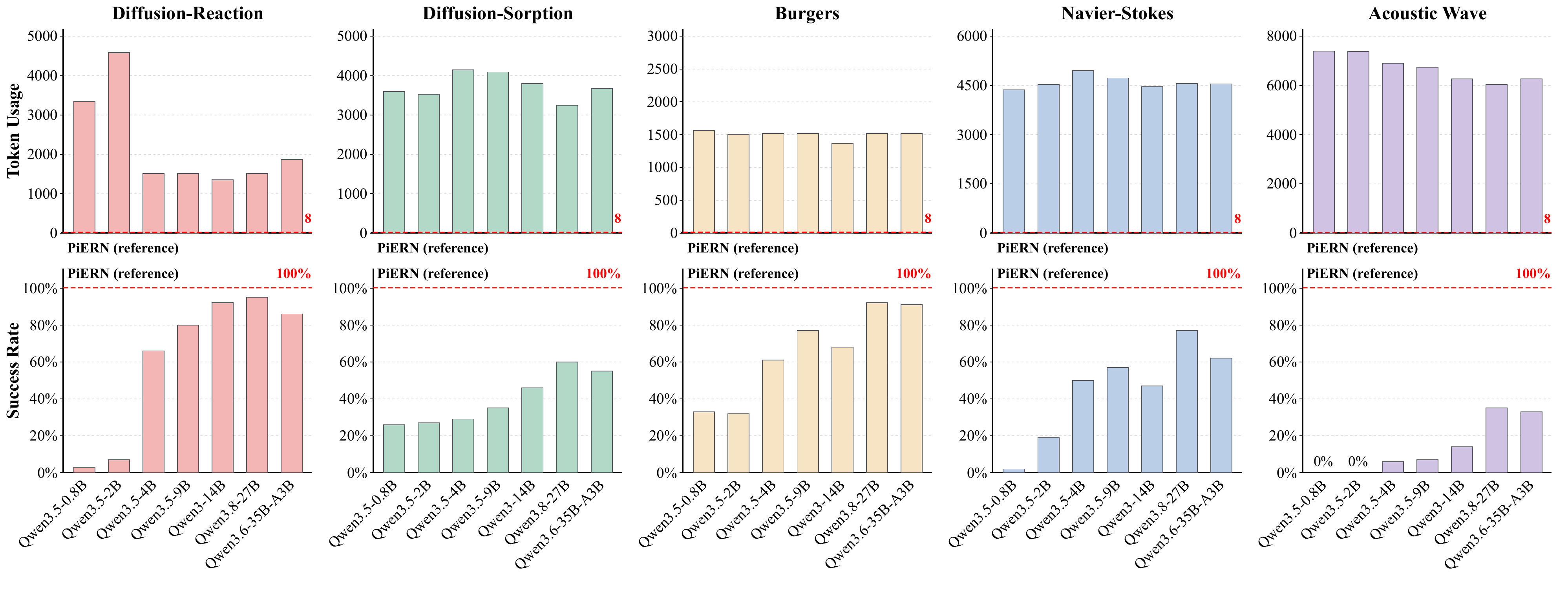}
    \caption{Token usage and success rate across different tasks comparing PiERN with MAF-based multi-agent systems. \textbf{The red dashed line denotes PiERN}.}
    \label{fig:token}
\end{figure}

Building on Microsoft Agent Framework (MAF), we constructed multi-agent systems using the same base LLMs of different sizes as PiERN. In these systems, different agents undertake specialized roles: LLMs are responsible for routing and dialogue interaction, while external experts are responsible for executing the corresponding tasks of high-precision numerical computation, and the agents collaborate through explicit communication. \begin{table}[!t]
\centering
\caption{Latency and GPU energy consumption of PiERN and MAF across five tasks.}
\label{tab:latency_gpu_comparison}
\newcolumntype{N}{>{\raggedleft\arraybackslash}X}
\newcommand{\taskname}[1]{\parbox[c][22pt][c]{\linewidth}{\centering\fontsize{7.5}{8.4}\selectfont\bfseries #1}}
\begingroup
\small
\setlength{\tabcolsep}{1.1pt}
\renewcommand{\arraystretch}{1.16}
\begin{adjustbox}{scale=0.8}
\begin{tabularx}{\linewidth}{@{}>{\raggedright\arraybackslash}p{72pt}*{5}{N}@{\hspace{6pt}}*{5}{N}@{}}
\toprule
\multirow{2}{72pt}[-7.5pt]{\textbf{Model (LLM)}}
 & \multicolumn{5}{c}{\textbf{Latency (s)}}
 & \multicolumn{5}{c}{\textbf{GPU Energy Consumption (kJ)}} \\
\cmidrule(lr){2-6}\cmidrule(lr){7-11}
 & \taskname{Diffusion\\Reaction} & \taskname{Diffusion\\Sorption} & \taskname{Burgers} & \taskname{Navier--\\Stokes} & \taskname{Acoustic\\Wave} & \taskname{Diffusion\\Reaction} & \taskname{Diffusion\\Sorption} & \taskname{Burgers} & \taskname{Navier--\\Stokes} & \taskname{Acoustic\\Wave} \\
\midrule
\multicolumn{11}{@{}l}{\textbf{\textit{PiERN}}} \\
Qwen3.5-0.8B & 0.62 & 0.63 & 0.62 & 0.64 & 0.77 & 0.056 & 0.078 & 0.057 & 0.123 & 0.171 \\
Qwen3.5-2B & 0.64 & 0.65 & 0.63 & 0.67 & 0.80 & 0.063 & 0.092 & 0.065 & 0.135 & 0.185 \\
Qwen3.5-4B & 0.74 & 0.75 & 0.74 & 0.82 & 0.98 & 0.094 & 0.109 & 0.096 & 0.170 & 0.229 \\
Qwen3.5-9B & 0.71 & 0.75 & 0.70 & 0.85 & 1.04 & 0.106 & 0.127 & 0.107 & 0.201 & 0.271 \\
Qwen3-14B & 0.77 & 0.81 & 0.77 & 0.97 & 1.20 & 0.135 & 0.162 & 0.138 & 0.257 & 0.350 \\
Qwen3.8-27B & 1.05 & 1.14 & 1.05 & 1.43 & 1.78 & 0.209 & 0.249 & 0.213 & 0.398 & 0.530 \\
Qwen3.6-35B-A3B & 1.19 & 1.25 & 1.22 & 1.28 & 1.44 & 0.142 & 0.166 & 0.151 & 0.215 & 0.282 \\
\midrule
\multicolumn{11}{@{}l}{\textbf{\textit{MAF}}} \\
Qwen3.5-0.8B & 58.85 & 42.30 & 10.70 & 46.26 & 108.06 & 5.921 & 4.249 & 1.122 & 4.746 & 11.917 \\
Qwen3.5-2B & 90.07 & 39.16 & 8.97 & 47.19 & 108.19 & 13.225 & 5.763 & 1.320 & 6.324 & 14.398 \\
Qwen3.5-4B & 12.50 & 74.06 & 12.35 & 83.06 & 135.82 & 1.933 & 11.592 & 1.911 & 12.667 & 20.728 \\
Qwen3.5-9B & 12.61 & 76.57 & 12.52 & 77.49 & 130.85 & 2.628 & 17.037 & 2.588 & 14.264 & 24.284 \\
Qwen3-14B & 10.14 & 65.36 & 10.51 & 59.73 & 101.03 & 2.697 & 18.102 & 2.756 & 15.141 & 25.317 \\
Qwen3.8-27B & 25.24 & 87.17 & 25.24 & 135.03 & 199.67 & 5.959 & 20.535 & 5.966 & 33.939 & 49.117 \\
Qwen3.6-35B-A3B & 42.14 & 98.09 & 21.00 & 114.18 & 179.55 & 6.597 & 15.274 & 3.284 & 17.743 & 29.140 \\
\bottomrule
\end{tabularx}
\end{adjustbox}
\par
\endgroup
\end{table}
 To highlight the performance advantages of the PiERN architecture, we design a series of comparative experiments to compare PiERN with these multi-agent systems. We compare performance along four dimensions, which directly reflect the core differences between PiERN and the multi-agent systems, with detailed information of these metrics provided in Appendix~\ref{app:metrics}. Results for LangGraph and Programmatic are also provided in Appendix~\ref{app:metrics}.

Table~\ref{tab:latency_gpu_comparison} presents the comparison of latency and GPU energy consumption across five computation-reasoning tasks for different models. The PiERN series models demonstrate overwhelming speed advantages across all tests, achieving latency reductions of 1 to 2 orders of magnitude compared to the MAF-based systems. Moreover, the PiERN architecture demonstrates exceptional energy efficiency, reducing energy consumption also by 1 to 2 orders. For instance, on Diffusion-Reaction with Qwen3.5-0.8B, PiERN consumes only 0.056~kJ, less than $1\%$ of the MAF-based system's energy consumption (5.921~kJ). The results indicate that PiERN significantly enhances computational efficiency while dramatically reducing computational costs.

Furthermore, as shown in Figure~\ref{fig:token}, PiERN consumes only 8 tokens across all five tasks, while the MAF-based systems typically require substantially more tokens to perform the same tasks. Moreover, despite consuming a large number of tokens, the MAF-based systems exhibit highly unstable expert invocation success rates, with some models nearing failure. In contrast, PiERN maintains a $100\%$ expert routing success rate. This comparison strongly demonstrates the efficiency and robustness of PiERN.

\begin{figure}[!hb]
    \centering
    \includegraphics[width=0.80\textwidth]{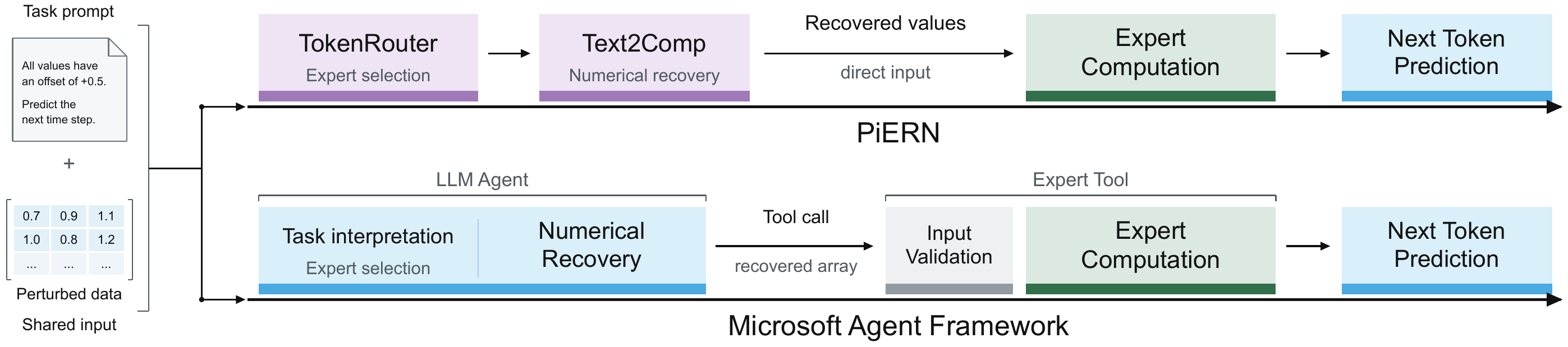}
    \caption{\textbf{Execution-flow decomposition for PiERN and the MAF-based system}.}
    \label{fig:token-decomp}
\end{figure}

Figure~\ref{fig:token-decomp} presents the execution-flow decomposition for PiERN and the MAF-based system during the execution of a single task. In PiERN, the inference process alternates between next-token prediction and high-precision expert computation, eliminating redundant overhead. In contrast, multi-agent systems must sequentially perform task analysis, data alignment, code generation, tool invocation, and result analysis, with each step introducing additional costs in terms of CoT \citep{wei2023chainofthoughtpromptingelicitsreasoning}. This decomposition clearly reveals why PiERN can achieve significantly faster inference.

\WFclear
\subsubsection{General Language Evaluation}
We evaluate the performance of PiERN by using lm-evaluation-harness \citep{eval-harness} framework, on the MMLU \citep{MMLU-benchmark} and GLUE \citep{GLUE-benchmark} benchmarks. As shown in \cref{tab:eval-results} (Appendix~\ref{app:general-language-evaluation}), the accuracy of PiERN on both MMLU and GLUE benchmarks closely matches that of base models, even as it introduces high-precision computation capabilities for scientific computation-reasoning tasks.
The chosen of GLUE and MMLU was based on their diverse language tasks.

\begin{figure}[!htbp]
    \centering
    \includegraphics[width=0.683443\textwidth]{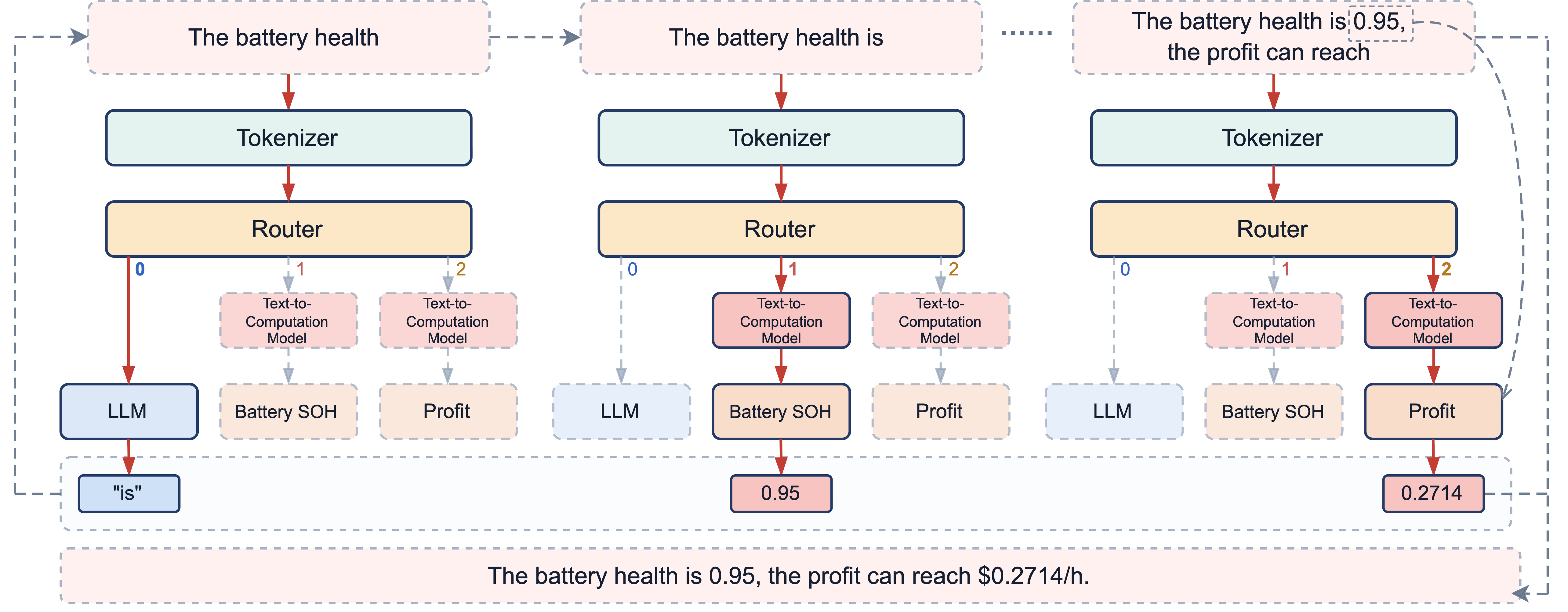}
    \caption{\textbf{PiERN-BMS Modeling Paradigm}. 
    \textbf{Left:} Routing to the LLM for reasoning. 
    \textbf{Middle:} Routing to a scientific expert for capacity estimation. 
    \textbf{Right:} Routing to another expert for profit computation \textbf{using the computational result from the previous expert}.}
    \label{fig:iterative-cal-architecture}
\end{figure}

\subsection{PiERN-BMS Modeling Paradigm}
To demonstrate PiERN's capability in coordinating multiple experts for collaborative iterative computations in complex tasks, we introduce the PiERN-BMS (Battery Management System) task. As shown in Figure~\ref{fig:iterative-cal-architecture}, PiERN-BMS integrates an LLM and two experts: a non-linear neural network expert for battery remaining capacity estimation, and a trained MLP expert for arbitrage profit prediction (More details are provided in Appendix~\ref{app:PiERN-BMS}). During the inference, the token router first invokes the non-linear expert to diagnose the health status; subsequently, the LLM utilizes the state computed by the previous expert and, through language-based reasoning, triggers the linear expert to perform the final profit computation. This iterative “text-computation-reasoning-computation” loop highlights that PiERN can also enable the collaborative iterative computation of outputs from different experts.
\section{Discussion}
In this study, we propose PiERN, an architecture that unifies high-precision experts’ computation with LLMs’ reasoning. PiERN goes beyond the traditional workflow paradigm of tool invocation by enabling token-level alternating execution of expert computation and language reasoning within a single CoT. It should be noted that PiERN demonstrates significant potential for large-scale agentic RL training \citep{gao2025rollartscalingagenticrl, zhang2025agentrlscalingagenticreinforcement}, particularly in terms of scaling along the time dimension. Moreover, the statistical paradigm of current LLMs is mainly embodied in inductive reasoning and analogical reasoning, while PiERN endogenously integrates high-precision computation (deductive reasoning) into LLMs. The integration of inductive, analogical, and deductive reasoning paradigms introduces a new paradigm of intelligence. Nevertheless, the current evaluation of PiERN still remains limited to the specific computation–reasoning tasks.

Future research will focus on three main directions: first, investigating compressed representations and exact reconstruction to enable scalable handling of high-dimensional inputs, including PiERN architecture which based on both LLMs as the numerical encoders and text-to-computation decoders \citep{Dosovitskiy2020AnII, 10.5555/3737916.3741776}; second, exploring automated methods for efficient alignment of large-scale domain knowledge, such as physical laws, operational protocols, and scheduling manuals, which encoded in textual form, to enable cross-modal text-to-computation fusion for scientific experts, thus overcoming the inefficiencies associated with manual data curation; finally, promoting the practical application of PiERN in complex system by incentivizing the computation-reasoning ability through reinforcement learning \citep{Guo2025}, such as power grid scheduling, drug discovery, and materials simulation, thereby verifying its feasibility and transformative potential as the infrastructure for next-generation scientific intelligence systems.

\section*{AI Use Statement}

In our experiments, we used LLMs to assist in implementing parts of the technical pipeline code. We have carefully reviewed and verified all generated codes. In preparing the manuscript, we used LLMs to translate parts of our drafts that had been carefully prepared, and to polish the language. All generated content has been thoroughly checked by us to ensure accuracy. We take full responsibility for the validity of the research results and the final content of the paper.

\section*{Impact Statement}

The PiERN architecture proposed in this study will significantly accelerate scientific research and engineering innovation. First, PiERN enables faster exploration of complex physical systems (e.g., fluid dynamics, power grid scheduling, meteorological prediction) through natural language interactions, with extremely low inference latency and high computational accuracy. Furthermore, PiERN's substantial improvements in inference speed and energy efficiency contribute to the advancement of "green AI", achieving a 1-2 order of magnitude increase in energy efficiency compared to the mainstream multi-agent systems, thereby greatly reducing computational resource consumption and carbon emissions. Through natural language interface, PiERN democratizes AI for Science technologies, further fostering cross-disciplinary innovations.

However, the potential negative impacts and ethical concerns of PiERN must also be acknowledged. First, excessive reliance on PiERN's "high-precision" outputs, without consideration of the applicability boundaries of the underlying expert models may introduce risks in sensitive domains such as aerospace safety or medical device design. Second, while PiERN's efficient physical simulation capabilities are beneficial for scientific research, they may also be misused to accelerate illicit research in sensitive areas (e.g., fluid dynamics simulations for controlled substances). To mitigate these risks, we recommend implementing cross-validation mechanisms in safety-critical applications. Overall, aside from the ethical considerations mentioned, no other significant societal negative impacts have been identified.

\bibliographystyle{plainnat}
\bibliography{ref}             
\clearpage
\appendix

\section{An End-to-End PiERN Running Example}
\label{app:PDEBench end-to-end runing example}

\subsection{PDE Task Descriptions}
\label{app:PDEBench_Task}

The 1d diffusion-reaction task involves solving a 1d PDE that models the diffusion and reaction of substances, covering both linear and nonlinear dynamics. The governing equation is
\begin{equation}
\frac{\partial u}{\partial t} = D \frac{\partial^2 u}{\partial x^2} + R(u),
\end{equation}
where \( u(x, t) \) represents the concentration of the substance at time \( t \) and position \( x \), \( D \) is the diffusion coefficient, and \( R(u) \) is the reaction term, which represents the chemical reaction rate. The 1d diffusion-sorption task builds on 1d diffusion-reaction task by incorporating adsorption phenomena, further increasing its complexity. The governing equation is
\begin{equation}
\frac{\partial u}{\partial t} = D \frac{\partial^2 u}{\partial x^2} - k_a u (1 - u),
\end{equation}
where \( u(x, t) \) is the concentration of the adsorbed substance, \( D \) is the diffusion coefficient, \( k_a \) is the adsorption constant, and \( (1 - u) \) represents the availability of adsorption sites. The Burgers equation is a widely used nonlinear PDE for modeling shock waves, turbulence, and other phenomena. The governing equation is
\begin{equation}
\frac{\partial u}{\partial t} + u \frac{\partial u}{\partial x} = \nu \frac{\partial^2 u}{\partial x^2},
\end{equation}
where \( u(x, t) \) is the velocity field, and \( \nu \) is the fluid viscosity.

The Navier-Stokes task models one-dimensional compressible fluid motion. The governing equations are~\citep{10.5555/3600270.3600387}
\begin{subequations}
\begin{gather}
\partial_t\rho+\nabla\cdot(\rho\mathbf{v}) = 0, \\
\rho(\partial_t\mathbf{v}+\mathbf{v}\cdot\nabla\mathbf{v}) = -\nabla p+\eta\nabla^2\mathbf{v}+(\zeta+\eta/3)\nabla(\nabla\cdot\mathbf{v}), \\
\partial_t\left(\epsilon+\frac{\rho|\mathbf{v}|^2}{2}\right)
+\nabla\cdot\left[\left(\epsilon+p+\frac{\rho|\mathbf{v}|^2}{2}\right)\mathbf{v}-\mathbf{v}\cdot\sigma'\right] = 0,
\end{gather}
\end{subequations}
where $\rho$ is the mass density, $\mathbf{v}$ is the fluid velocity, $p$ is the gas pressure, $\epsilon=p/(\Gamma-1)$ is the internal energy density with adiabatic index $\Gamma=5/3$, $\sigma'$ is the viscous stress tensor, and $\eta$ and $\zeta$ are the shear and bulk viscosities, respectively. The Acoustic Wave task models acoustic-wave propagation in a two-dimensional medium with constant density and spatially varying wave speed. The governing equation is~\citep{deng2021openfwi}
\begin{equation}
\nabla^2p(\mathbf{r},t)-\frac{1}{c(\mathbf{r})^2}\frac{\partial^2p(\mathbf{r},t)}{\partial t^2}=s(\mathbf{r},t),
\end{equation}
where $\mathbf{r}=(x,z)$ denotes spatial position, $t$ is time, $p(\mathbf{r},t)$ is the pressure wavefield, $c(\mathbf{r})$ is the wave speed, and $s(\mathbf{r},t)$ is the seismic source term.

\paragraph{Data Preprocessing.}
For the PDEBench tasks, the raw data are derived from the official PDEBench. To reduce computational cost, we apply fixed-step spatial downsampling (reducing resolution from 1024 to 64 or 32 grid points) and temporal sliding windows to generate historical-future state pairs $(u_t, u_{t+1})$. These downsampled numerical pairs serve as the numerical ground truth for our subsequent computation-reasoning tasks data synthesis pipeline.

\subsection{Language Templates and Data Synthesis}
\label{app:data_formulation}
\begin{figure}[!ht] 
  \centering
  \includegraphics[width=\linewidth]{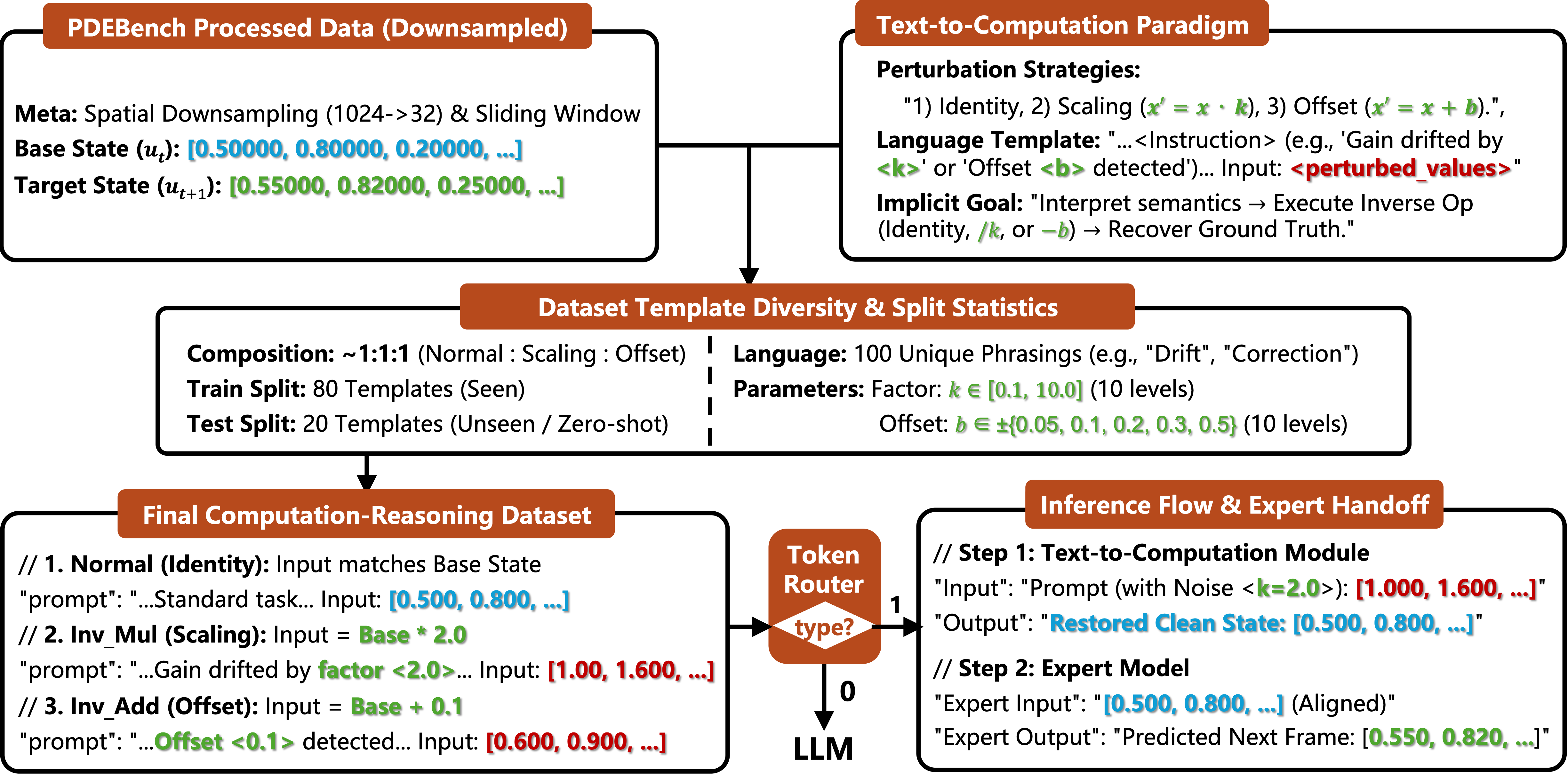}
  \vspace{-0.2cm}
  \caption{\textbf{The language templates and data synthesis pipeline.} 
  \textbf{Top:} Raw data is spatially downsampled and paired, then injected with controlled perturbations (e.g., Scaling $\times k$) via natural language templates. 
\textbf{Middle:} An 80/20 split to evaluate text-to-computation module generalization. 
\textbf{Bottom:} Token router dynamically chose the LLM or experts.
}
  \label{fig:synthesis_pipeline}
  
\end{figure}

As shown in Figure~\ref{fig:synthesis_pipeline}, to endow PiERN with the ability to efficiently integrate language text with numerical computation and perform token-level routing, we constructed a \textbf{language templates and data synthesis pipeline}, which transforms raw numerical pairs into pairs that combine both text and numbers, and we further constructed the data for training the token router.

\subsubsection{Text-to-Computation Language Patterns}
We designed \textbf{$N=100$ unique language patterns} to wrap the numerical data pairs within natural language contexts.
\begin{itemize}
    \item \textbf{Perturbation Strategies:} The dataset is balanced across three types of transformations, which are applied to the numerical pairs through corresponding textual description, and with an approximate ratio of $1:1:1$:
    \begin{itemize}
        \item \textbf{Identity:} Standard prediction tasks where the model learns to predict values directly from the input.
        \item \textbf{Scaling:} The input $x$ is multiplied by a constant factor $k$ ($x' = x \cdot k$), and the model must learn to divide by $k$.
        \item \textbf{Offset:} The input $x$ is shifted by a constant $b$ ($x' = x + b$), and the model must learn to subtract $b$.
    \end{itemize}
    \item \textbf{Split Strategy:} We enforce an \textbf{80/20 split}, reserving 20 templates exclusively for testing (Zero-shot). This evaluates the PiERN’s ability to generalize to unseen test language instruction patterns, rather than memorizing.
\end{itemize}

\subsubsection{PiERN Training Data Instances}
To clarify the training process, we present concrete examples of token router and text-to-computation module of PiERN from our dataset. These examples illustrate the specific input-output pairs for training.

\begin{tcolorbox}[colback=gray!5!white, colframe=gray!75!black, title=\textbf{Example 1: Training Sample for Text-to-Computation Module}]
\textbf{Task Type:} Offset (Correction of System Error $b=0.1$)

\vspace{0.2cm}
\textbf{Input Prompt (Text + Numbers):} \\
\textit{``The following is the history record of the 1d diffusion-reaction state. The data contains a system error of 0.1. Please subtract 0.1 to correct it and use the FNO model to calculate the next frame. Input stream: [0.79920, 0.79920, 0.79920, ...]''}

\vspace{0.2cm}
\textbf{Target Output (Reconstruction Tensor):} \\
\texttt{[0.69920, 0.69920, 0.69920, ...]}

\vspace{0.2cm}
\small{\textit{Note: The text-to-computation module learns to execute the operation $x_{clean} = x_{input} - 0.1$ as instructed by the semantic cue “subtract 0.1”,thereby allowing the potential text semantics to influence the numerical input of the expert, enabling text-numeric multimodal learning..}}
\end{tcolorbox}

\begin{tcolorbox}[colback=gray!5!white, colframe=gray!75!black, title=\textbf{Example 2: Training Sample for Token Router}]
The token router is trained to output \texttt{Type 1} (Computation Mode) \textbf{only} when the model response is fully prepared for expert high-precision computation.

\vspace{0.2cm}
\textbf{Case A: High-Precision Computation Trigger (Type 1)} \\
\textbf{Input Context:} \textit{``...The data contains a zero-point drift (offset 0.1). Please subtract 0.1 compensation... Data follows: [0.80836, ...]. Okay, the scientific computation prediction result is:''} \\
\textbf{Label:} \textbf{1 (Trigger Expert for High-Precision Computation)} 

\vspace{0.2cm}
\hrule
\vspace{0.2cm}

\textbf{Case B: Ongoing Next Token Prediction Generation (Type 0)} \\
\textbf{Input Context:} \textit{``Okay, the scientific computation prediction''} \\
\textbf{Label:} \textbf{0 (Continue Next Token Prediction)}
\end{tcolorbox}

\subsection{An End-to-End PiERN Inference Running Example}
During inference, PiERN dynamically passes data between the LLM and Experts. The following workflow illustrates the core mechanism for the 1d diffusion-reaction task described above (Offset $b=0.1$).

\textbf{User Prompt:} \textit{``The data contains a system error of 0.1. Please subtract 0.1... Input stream: [0.79920, ...]''}

\begin{enumerate}
    \item \textbf{Phase 1: Semantic Parsing \& LLM Reasoning Phase} \\
    The LLM generates/processes the prefix text. The \textbf{Token Router} monitors each token (Output \texttt{Type 0}) until the instruction is fully formulated.

    \item \textbf{Phase 2: Token Router Trigger} \\
    \textbf{Trigger:} The LLM generates the final cue: \textit{``...prediction result is:''}. \\
    \textbf{Action:} The Router detects this specific suffix and switches output to \textbf{Type 1}. \\
    \textbf{Transfer:} Text generation is halted. The full context (instruction + noisy data \texttt{[0.79920...]...prediction result is:}) is passed to the \textbf{Text-to-Computation Module}.

    \item \textbf{Phase 3: Reconstruction \& Executing Experts High-Precision Computation} \\
    \textbf{Restoration:} The module interprets \textit{``subtract 0.1''} and executes the correction:
    \begin{equation*}
        \hat{x}_{clean} = [0.79920, \dots] - 0.1 = [0.69920, \dots]
    \end{equation*}
    \textbf{Expert Execution:} The reconstruction tensor $\hat{x}_{clean}$ is then forwarded to the frozen 1d diffusion-reaction expert. \\
    \textbf{Prediction:} The expert calculates the next state: $y_{pred} = [0.72550, \dots]$.

    \item \textbf{Phase 4: Response Output} \\
    The numerical result $y_{pred}$ is appended to the response. \\
    \textbf{Final Output:} \textit{``...prediction result is: [0.72550, 0.72550, ...].''}
\end{enumerate}

\par
\newpage
\section{Experimental Details for Component Generalization}
\label{app:component-generalization}

This appendix provides the experimental settings for the component generalization study in Section~\ref{sec:component-generalization}.

\textbf{Task and template protocol.} We evaluate the text-to-computation module (text2comp) and Token Router separately on Diffusion-Sorption, using Identity, Scaling, and Offset instructions. Seen and unseen requests are paired on the same test numerical inputs, isolating the effect of instruction-template changes. Both components are based on Qwen3.5-0.8B and trained independently.

\textbf{text2comp evaluation.} We evaluate 1,000 template pairs with 100 numerical samples per pair, yielding 100,000 requests per split. The metric is mean input reconstruction RMSE.

\textbf{Router evaluation.} The Router is evaluated separately on 1,000 template pairs with 1,000 numerical samples per pair, yielding 1,000,000 requests per split. At predefined token decision points, it predicts whether to continue language generation or invoke the task's expert. Accuracy is measured over these decision points, totaling 22,000,000 across both splits.

\section{Metrics for PiERN and Multi-Agent Systems}
\label{app:metrics}
We compare performance along four dimensions, which directly reflect the core differences between PiERN and the multi-agent systems: \textbf{(i) Latency}: determines the response speed of the system in interactive scenarios, directly affecting user experience and the feasibility of real-time decision-making tasks;  
\textbf{(ii) Token Usage}: measures the additional overhead in inference caused by long-context understanding and cross-agent communication, directly reflecting the user-side inference cost and the overall economic efficiency of task execution;  
\textbf{(iii) GPU Energy Consumption}: reflects resource utilization and energy efficiency, serving as a key metric for evaluating deployment scalability and sustainability;
\textbf{(iv) Success Rate}: measures the proportion of correct high-precision results obtained in computation–reasoning tasks, directly reflecting the system’s reliability and task completion capability.

\begin{figure}[H]
    \centering
    \includegraphics[width=0.797031\textwidth]{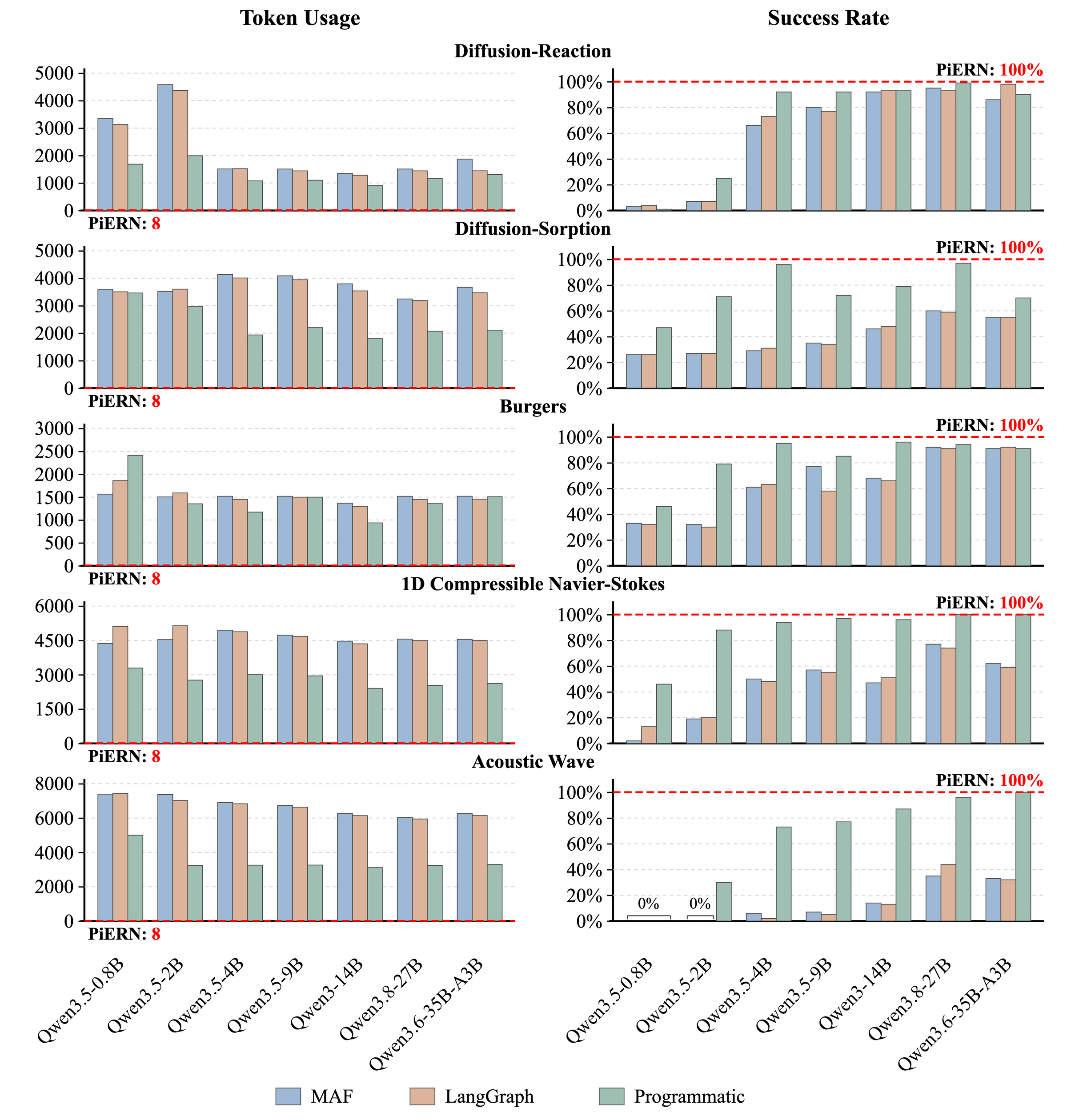}
    \caption{Token usage and success rate across different tasks comparing  PiERN with multi-agent systems. \textbf{The red dashed line denotes PiERN}.}
    \label{fig:token-all-frameworks}
\end{figure}

Table~\ref{tab:latency_gpu_appendix} reports the latency and GPU energy consumption of LangGraph and Programmatic across the five tasks. Figure~\ref{fig:token-all-frameworks} compares token usage and success rate across MAF, LangGraph, and Programmatic, with PiERN as the reference.
\begin{table}[H]
\centering
\caption{Latency and GPU energy consumption of LangGraph and Programmatic across five tasks.}
\label{tab:latency_gpu_appendix}
\newcolumntype{V}{>{\raggedleft\arraybackslash}X}
\newcommand{\estimated}[1]{#1}
\newcommand{\appendixtaskname}[1]{\parbox[c][22pt][c]{\linewidth}{\centering\fontsize{7.5}{8.4}\selectfont\bfseries #1}}
\begingroup
\small
\setlength{\tabcolsep}{1.1pt}
\renewcommand{\arraystretch}{1.16}
\begin{tabularx}{\linewidth}{@{}>{\raggedright\arraybackslash}p{72pt}*{5}{V}@{\hspace{6pt}}*{5}{V}@{}}
\toprule
\multirow{2}{72pt}[-7.5pt]{\textbf{Model (LLM)}}
 & \multicolumn{5}{c}{\textbf{Latency (s)}}
 & \multicolumn{5}{c}{\textbf{GPU Energy Consumption (kJ)}} \\
\cmidrule(lr){2-6}\cmidrule(lr){7-11}
 & \appendixtaskname{Diffusion\\Reaction} & \appendixtaskname{Diffusion\\Sorption} & \appendixtaskname{Burgers} & \appendixtaskname{Navier--\\Stokes} & \appendixtaskname{Acoustic\\Wave} & \appendixtaskname{Diffusion\\Reaction} & \appendixtaskname{Diffusion\\Sorption} & \appendixtaskname{Burgers} & \appendixtaskname{Navier--\\Stokes} & \appendixtaskname{Acoustic\\Wave} \\
\midrule
\multicolumn{11}{@{}l}{\textbf{\textit{LangGraph}}} \\
Qwen3.5-0.8B & 54.96 & 41.73 & 20.42 & 65.68 & 108.94 & 5.511 & 4.187 & 2.076 & 7.240 & 13.660 \\
Qwen3.5-2B & 86.23 & 43.30 & 12.81 & 65.81 & 101.59 & 12.658 & 6.358 & 1.888 & 8.787 & 13.254 \\
Qwen3.5-4B & 15.23 & 72.24 & 12.35 & 85.08 & 135.01 & 2.354 & 11.208 & 1.911 & 12.889 & 20.738 \\
Qwen3.5-9B & 12.54 & 70.72 & 14.14 & 78.13 & 123.84 & 2.581 & 14.648 & 2.944 & 14.552 & 23.387 \\
Qwen3-14B & 10.20 & 53.45 & 10.45 & 62.39 & 101.57 & 2.692 & 14.411 & 2.737 & 15.114 & 25.100 \\
Qwen3.8-27B & 26.22 & 96.73 & 25.48 & 133.75 & 194.89 & 6.053 & 21.628 & 6.030 & 32.926 & 46.974 \\
Qwen3.6-35B-A3B & 24.25 & 98.94 & 23.18 & 114.58 & 176.81 & 3.641 & 14.978 & 3.527 & 18.040 & 26.415 \\
\midrule
\multicolumn{11}{@{}l}{\textbf{\textit{Programmatic}}} \\
Qwen3.5-0.8B & 19.99 & 44.42 & 39.15 & 23.81 & 52.83 & 2.034 & 4.487 & 3.935 & 2.677 & 5.506 \\
Qwen3.5-2B & 27.39 & 31.21 & 10.49 & 9.79 & 2.86 & 4.027 & 4.568 & 1.536 & 1.303 & 0.409 \\
Qwen3.5-4B & 4.68 & 4.18 & 7.59 & 22.28 & 4.37 & 0.726 & \estimated{0.668} & 1.176 & 3.427 & 0.736 \\
Qwen3.5-9B & 5.55 & 13.85 & 19.29 & 20.22 & 4.96 & 1.143 & \estimated{2.881} & 3.961 & 3.789 & 0.997 \\
Qwen3-14B & 2.94 & 3.35 & 3.20 & 3.75 & 3.65 & 0.768 & \estimated{0.872} & 0.815 & 0.934 & 0.951 \\
Qwen3.8-27B & 17.03 & 19.44 & 31.35 & 8.81 & 8.33 & 3.793 & \estimated{4.516} & 6.978 & 2.373 & 2.277 \\
Qwen3.6-35B-A3B & 22.59 & 17.34 & 34.17 & 13.04 & 10.38 & 3.490 & \estimated{2.740} & 5.258 & 2.070 & 1.671 \\
\bottomrule
\end{tabularx}
\par
\endgroup
\end{table}

\section{General Language Evaluation}
\label{app:general-language-evaluation}

\begin{table}[H]
\centering
\caption{Performance Comparison of PiERN and Qwen Models on MMLU and GLUE Benchmark}
\resizebox{\textwidth}{!}{
\begin{tabular}{l|ccccccc}
\toprule
\textbf{Model} & \textbf{MMLU} & \textbf{MNLI} & \textbf{RTE} & \textbf{SST2} & \textbf{QQP} & \textbf{QNLI} & \textbf{MRPC} \\
\midrule
\centering PiERN-2.4B & 50.14 & 33.50 & 53.09 & 51.43 & 36.87 & 49.65 & 67.83 \\
\centering Qwen3.5-0.8B & 50.24 & 33.25 & 52.35 & 51.03 & 36.81 & 49.55 & 68.38 \\
\midrule
\centering PiERN-3.6B & 57.67 & 46.57 & 64.87 & 80.80 & 78.59 & 62.08 & 74.11 \\
\centering Qwen3.5-2B & 57.62 & 46.58 & 64.26 & 80.16 & 79.90 & 62.02 & 72.79 \\
\midrule
\centering PiERN-5.6B & 70.23 & 47.04 & 73.30 & 89.00 & 64.41 & 76.45 & 47.54 \\
\centering Qwen3.5-4B & 70.12 & 46.77 & 73.29 & 88.88 & 64.49 & 76.37 & 47.30 \\
\midrule
\centering PiERN-10.6B & 70.70 & 39.98 & 81.58 & 89.46 & 83.56 & 54.33 & 75.17 \\
\centering Qwen3.5-9B & 70.64 & 39.61 & 81.59 & 89.45 & 83.42 & 54.59 & 75.00 \\
\midrule
\centering PiERN-15.6B & 78.93 & 67.32 & 77.64 & 92.42 & 82.34 & 83.75 & 78.42 \\
\centering Qwen3-14B & 78.82 & 67.02 & 77.62 & 92.43 & 82.29 & 83.96 & 78.68 \\
\midrule
\centering PiERN-28.6B & 80.27 & 64.33 & 83.20 & 87.92 & 73.70 & 83.37 & 41.17 \\
\centering Qwen3.8-27B & 80.22 & 64.27 & 82.67 & 88.07 & 73.81 & 83.49 & 40.69 \\
\midrule
\centering PiERN-36.6B & 82.76 & 60.05 & 76.88 & 87.50 & 80.93 & 80.17 & 78.93 \\
\centering \makebox[103.61pt][l]{Qwen3.6-35B-A3B} & 82.77 & 60.05 & 76.17 & 87.73 & 81.06 & 80.27 & 78.19 \\
\bottomrule
\end{tabular}
  } \label{tab:eval-results}

\end{table}

\section{PiERN-BMS Modeling Paradigm}
\label{app:PiERN-BMS}

\begin{figure}[!htbp]
    \centering
    \includegraphics[width=0.80\textwidth]{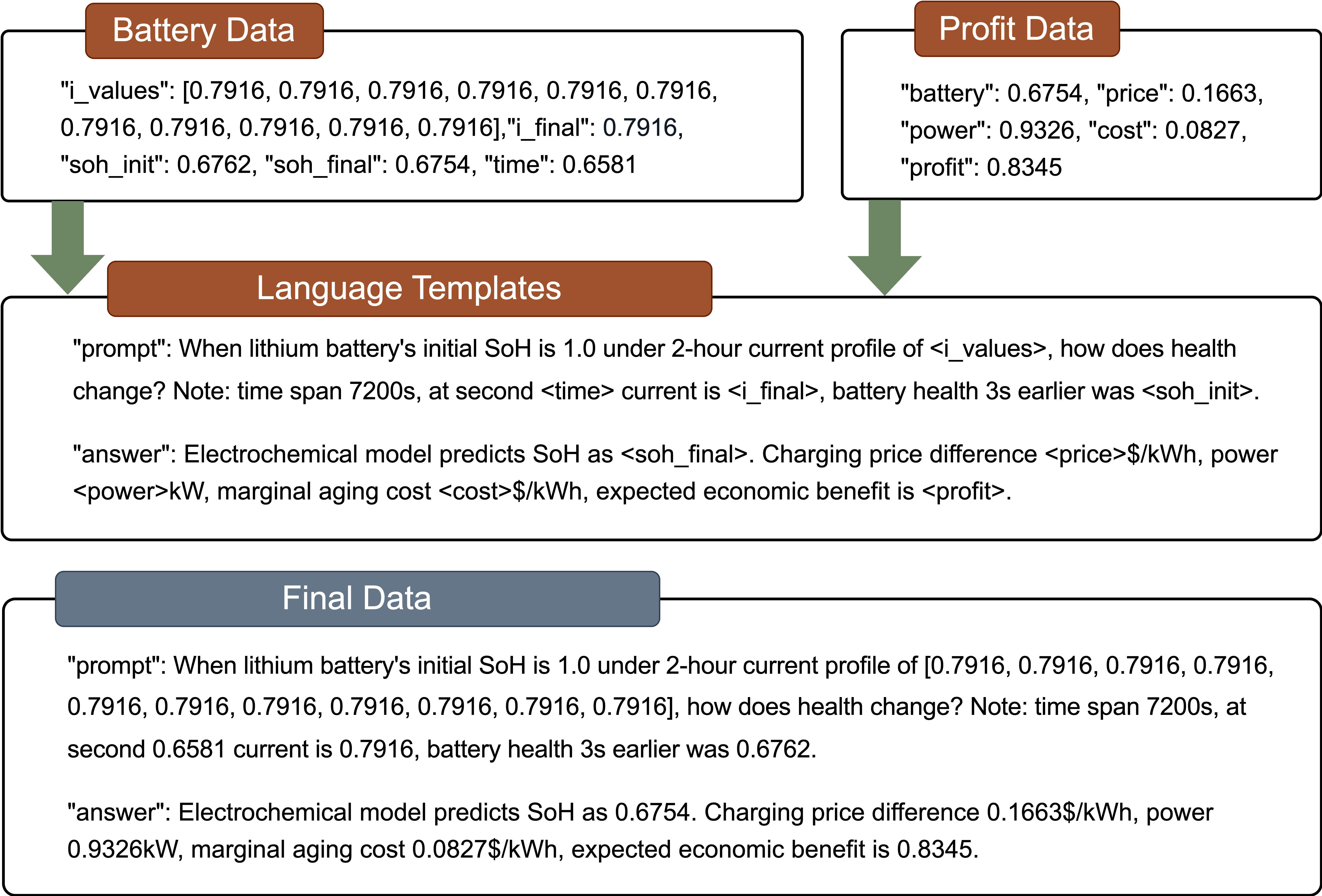}
    \caption{\textbf{Pipeline of language-template-based data synthesis for PiERN-BMS}. \textbf{Top}: Extraction of raw physical and numerical
information, including battery capacity data and profit data. \textbf{Middle}: Application of Language templates to wrap structured data into
natural language contexts. \textbf{Bottom}: The Final data stream, where reasoning tokens and computational tokens are interleaved to form a
unified training sequence.}
    \label{fig:soh-data}
\end{figure}

In this section, we provide a detailed description of the PiERN-BMS (Battery Management System). We describe the battery remaining capacity estimation and arbitrage profit prediction tasks, implemented with DeepONet and MLP experts, respectively, along with the data synthesis process used to construct the dataset for PiERN-BMS.

\subsection{Battery Remaining Capacity Prediction Task (Non-Linear Task)}
\label{app:non-linear_task_data_description}

The primary objective of this task is to predict the remaining State of Health (SoH) of a battery based on time-series current data. Battery health is a core indicator for measuring battery performance degradation, typically defined as the ratio of the current remaining capacity to the initial capacity of the battery:
\begin{equation}
    SoH = \frac{C_{current}}{C_{initial}} \times 100\%
\end{equation}
The battery degradation is a complex non-linear process affected by multiple coupled physicochemical reactions and mechanisms, such as electrode material aging, electrolyte decomposition, and solid electrolyte interphase growth. These dynamics are usually modelled with PDEs. In the PiERN-BMS framework, this task serves as the benchmark for the non-linear neural network expert.

The input data for this task comprises two main categories of information constituting 13 input feature values. The first component is the time-series current data, consisting of 11 current values collected over a 2-hour period with a sampling interval of 12 minutes. The second component includes the specific time point to be predicted and its corresponding current value. The initial health status of the battery is set to 1 by default.

Regarding the data scale and source, the dataset is generated based on the Pseudo-Two-Dimensional (P2D) model constructed via COMSOL Multiphysics modeling. The repository contains a total of 9,600 samples matching the pattern of "current data - time point - state of health". Specifically, the training dataset contains 7,200 samples, while the test dataset contains 2,400 samples. 

\subsection{Battery Arbitrage Profit Calculation Task (Linear Task)}
\label{app:linear_task_data_description}

The core of this task is to calculate the battery profit from arbitrage in electricity markets or electricity bill saving. This represents a linear calculation task where the model must perform precise arithmetic based on physical states and market conditions. The calculation involves four key parameters: the battery degradation coefficient $\alpha$, the price difference between charging and discharging $\Delta p$, the battery charge-discharge power $P$, and the marginal degradation cost $c_a$. The final profit $R$ is formulated as:
\begin{equation}
    R = \Delta p \cdot P - \alpha \cdot c_a \cdot 1200
\end{equation}
where 1200 represents a specific time constant or cycle factor.

The dataset for this task consists of 10,000 data entries. In terms of data partitioning, the training data accounts for 90\% of the total dataset, with the remaining 10\% allocated for testing. 

\subsection{PiERN-BMS Modeling Paradigm}
\label{app:language_templates}

As shown in Figure~\ref{fig:soh-data}, to explore the collaborative iterative computation capabilities of the PiERN architecture within multiple experts and to better train the text-to-computation module and token-level router, we have designed multiple language templates specifically for the two tasks described above. These templates are essential for bridging the gap between raw numerical data and natural language reasoning. By combining the templates with the numeric data from the P2D model and market simulations, we enable the text-to-computation module to understand semantic contexts and generate numbers accurately across various scenarios.

The data construction process involves wrapping the 13-feature vector of the non-linear task and the parameters of the linear task into a cohesive narrative. For instance, a data sample starts with a user query embedding the current profile, prompting the router to trigger the non-linear expert for SoH estimation. This is followed by an LLM reasoning phase where the health state is analyzed, leading to a decision phase where market parameters are introduced. Finally, the router triggers the profit-prediction expert to compute the profit. This interleaved format of “text-computation-reasoning-computation” ensures the model learns to coordinate different experts effectively.

\subsection{Performance over LLM finetuning} 
Building upon the constructed PiERN-BMS modeling paradigm and datasets, we further conduct experiments to evaluate the performance of PiERN against traditional LLM finetuning approaches.

\textbf{Setups.} Finetuning is a common solution to bring domain knowledge into LLMs, thereby it can also be used to enhance LLMs with high-precision computation capabilities. We compare the MSE among the finetuned LLMs and PiERN on the test data of both Non-Linear Task and Linear Task. For fair comparison, LLMs are finetuned by the same training data used in the training of expert models. Meanwhile, only one language template is used to generate training data to let LLMs focus on computation. On the selection of LLMs, we used open-source Qwen3.5 models of various sizes (0.8B, 2B, 4B, and 9B) with full-parameter finetuning, while PiERN uses Qwen3.5-0.8B for its text-to-computation modules.

\begin{table}[!htbp]
    \centering
    \caption{Comparison of MSE on battery tasks.}
    \renewcommand{\arraystretch}{1.2}
    \setlength{\tabcolsep}{4pt}
        \begin{tabular*}{\textwidth}{@{\extracolsep{\fill}}l|c|cccc@{}}
            \hline
            \multirow{2}{*}{\textbf{Task}} &
            \multirow{2}{*}{\makecell{\textbf{PiERN} \\ \textbf{(Ours)}}} &
            \multicolumn{4}{c}{\textbf{Qwen3.5 (SFT)}} \\
            \cline{3-6}
            & & \textbf{0.8B} & \textbf{2B} & \textbf{4B} & \textbf{9B} \\
            \hline
            Non-Linear Task & $\mathbf{9.095\times10^{-5}}$ & $3.098\times10^{-2}$ & $2.818\times10^{-3}$ & $9.272\times10^{-3}$ & $4.357\times10^{-3}$ \\
            \hline
            Linear Task & $\mathbf{1.772\times10^{-5}}$ & $1.243\times10^{-4}$ & $1.056\times10^{-4}$ & $2.567\times10^{-5}$ & $2.303\times10^{-5}$ \\
            \hline
        \end{tabular*}
    \label{tab:mse_results_grouped_centered_vertical}  
\end{table}

\textbf{Results.} Table~\ref{tab:mse_results_grouped_centered_vertical} shows the accuracy of PiERN on these two tasks is consistently better than all finetuned LLMs. Our method has the lowest MSE in all cases. The MSE of PiERN on the Non-Linear Task is one to two orders of magnitude lower than that of the best finetuned baseline.

Due to the pre-training data, LLMs are better at text understanding and generation than computational tasks. Therefore, LLMs usually cannot achieve high accuracy in computation tasks even after finetuning. Moreover, because the training of LLMs is an end-to-end, data-driven process, the models have very low interpretability. By comparison, our method integrates a pre-trained expert model (DeepONet for SoH prediction and MLP for profit prediction), which greatly enhances the model interpretability and stability. In summary, our method has demonstrated its advantages over other LLMs even with finetuning in terms of accuracy and interpretability in computation-reasoning tasks.
\end{document}